\documentclass{article}

    \usepackage[final, nonatbib]{neurips_data_2023}

\usepackage[utf8]{inputenc} 
\usepackage[T1]{fontenc}    
\usepackage{hyperref}       
\usepackage{url}            
\usepackage{booktabs}       
\usepackage{amsfonts}       
\usepackage{nicefrac}       
\usepackage{microtype}      
\usepackage{xcolor}         
\usepackage{makecell,tabularx}
\newcolumntype{L}{>{$}l<{$}}
\newcolumntype{C}{>{$}c<{$}}
\definecolor{mygray}{gray}{0.55}
\usepackage{graphicx}
\usepackage{subcaption}
\usepackage{amsmath} 
\usepackage{diagbox}
\usepackage{multirow}
\usepackage{bbm}

\title{On the Truthfulness of `Surprisingly Likely' Responses of Large Language Models}

\author{%
Naman Goel \\
University of Oxford\\
\texttt{naman.goel@cs.ox.ac.uk}\\
}

\begin{document}

\maketitle
\begin{abstract}
The principle of rewarding a crowd for surprisingly common answers has been used in the literature for designing a number of truthful information elicitation mechanisms~\cite{prelec2004bayesian, dasgupta2013crowdsourced, shnayder2016informed, radanovic2016incentives, kong2019information, faltings2022game, ijcai2023p740}. A related method has also been proposed in the literature for better aggregation of crowd wisdom~\cite{prelec2017solution}. Drawing a comparison between crowd based collective intelligence systems and large language models, we define the notion of `surprisingly likely' textual response of a large language model. This notion is inspired by the surprisingly common principle, but tailored for text in a language model. Using benchmarks such as TruthfulQA and openly available LLMs: GPT-2 and LLaMA-2, we show that the surprisingly likely textual responses of large language models are more accurate in many cases compared to standard baselines. For example, we observe up to 24 percentage points aggregate improvement on TruthfulQA and up to 70 percentage points improvement on individual categories of questions in this benchmark. We also provide further analysis of the results, including the cases when surprisingly likely responses are less or not more accurate.
\end{abstract}

\section{Introduction}\label{sec:intro}
Recent demonstrations of the promising capabilities of large language models (LLMs) have raised hopes about their successful deployment in a wide range of useful applications. However, one major issue with state-of-the-art LLMs that casts doubts on this optimism is their tendency to generate factually incorrect text. There are various ongoing efforts to address this issue. Promising efforts include scaling~\cite{wei2022emergent}, retrieval augmentation/grounding~\cite{lewis2020retrieval}, in-context learning~\cite{brown2020language}, chain-of-thought reasoning~\cite{wei2022chain} and other prompting, self-consistency~\cite{wang2023selfconsistency}, advanced decoding techniques~\cite{chuang2023dola}, uncertainty measurement based techniques~\cite{xiao2021hallucination}, self-improvement~\cite{pan2023automatically}, shifting model activations during inference~\cite{li2023inferencetime}, locating and editing factual information~\cite{meng2022locating}, learning from human feedback~\cite{ouyang2022training}, learning from AI feedback~\cite{bai2022constitutional, lee2023rlaif, shinn2023reflexion, weng2022large}, etc. While very useful, none of these efforts have fully solved the problem. Many scholars believe that the LLM architectures are fundamentally limited~\cite{marcus2020next,lecuntweet, xu2024hallucination} and thus, cannot be guaranteed to always generate factually correct text. Despite this, it remains an interesting research direction to understand the various sources of this multi-faceted problem~\cite{ye2023cognitive}, and the strengths and weaknesses of various complementary approaches for improvement. 

It is plausible that in some cases, LLMs generate incorrect information due to non-truthful objective. LLMs are commonly trained to generate answers that have the maximum probability conditioned on the prompt~\cite{borealisai}. The probability depends on the data points in its noisy training set. Given various data points in the training set, an LLM may generate correct or incorrect answers. An interesting demonstration is due to Lin \textit{et al.}, 2022~\cite{lin-etal-2022-truthfulqa}, who showed that LLMs generated false answers that mimic popular misconceptions from human texts. For example, when asked ``Which forbidden fruit Adam and Eve eat in the Garden of Eden?'', an LLM may answer that Adam and Eve eat an apple, which is a popular misconception; the correct answer is that the Bible does not specify the fruit. In this paper, we investigate whether and to what extent, the surprisingly common principle from the game-theoretic truthful information elicitation literature can be useful in avoiding this behaviour. 

There is a significant literature on game-theoretic incentive mechanisms for eliciting information from a crowd of agents. The mechanisms are also referred to as the peer-prediction or peer-consistency mechanisms~\cite{ijcai2023p740}. A seminal contribution in this literature is due to Prelec, 2004~\cite{prelec2004bayesian}. Prelec designs a mechanism (called the Bayesian Truth Serum) such that an agent answering a question, can maximize its expected incentive score by telling what the agent believes to be the correct answer, instead of telling what it believes most of the other agents would tell. The key idea in the Bayesian Truth Serum (BTS) is to reward surprisingly common answer (i.e. the answer that is more commonly reported than predicted a priori by agents, instead of the answer that is merely the most commonly reported one). Interesting examples, where BTS is particularly useful, include eliciting objective information that is rare or difficult to obtain and subjective information (e.g. subjective opinions) that is impossible to verify. Theoretical guarantees apply more generally beyond these examples. Besides incentive-compatibility
, it has also been shown that an information aggregation method based on the surprisingly common principle can be used to select more correct answers in crowdsourcing~\cite{prelec2017solution}. Further research in this area used motivation from the surprisingly common principle of the BTS to design mechanisms for various crowdsourcing settings. While the BTS asks agents to submit two reports for a question to determine which answers are surprisingly common, the ``minimal'' mechanisms proposed later in the literature, ask agents to submit only one. We discuss this in further detail in Section~\ref{sec:related}.

An LLM is very different from a crowd of agents in many ways. For example, the agents in crowdsourcing literature are assumed to be able to make independent observations about the world by incurring variable cost, form and update beliefs, and strategically (mis-)report their beliefs to maximize the incentives. On the other hand, the same cannot be said about a a pre-trained language model. Therefore, the theory of incentive mechanisms for crowd does not apply verbatim to an LLM. However, as we will discuss in Section~\ref{sec:discussion-pts-llm}, a comparison between crowdsourcing and how an LLM ``aggregates'' information from its training data points and generates answers when prompted, motivates us to take a step towards exploring the connections between the two fields of research. In this paper, we restrict the discussion on ``truth'' to objective information that can be clearly categorized as correct or incorrect (please see Section~\ref{sec:limitations} for critical discussion on the scope of this work). 

We define the notion of `surprisingly likely' textual response of a large language model. It is inspired from the surprisingly common principle developed in the information elicitation literature. In particular, we draw inspiration from the peer-truth serum mechanism~\cite{radanovic2016incentives}. While the peer-truth serum and other related mechanisms were developed for eliciting numerical or categorical answers from a crowd of agents, our `surprisingly likely' measure is adapted for textual responses of a single LLM that has been pre-trained on data from various sources. Through experiments on TruthfulQA, COPA and StoryCloze benchmarks, we show that surprisingly likely answers in large language models are indeed more correct in several cases. For example, on the TruthfulQA benchmark, we find significant gains in accuracy across different LLMs (up to 24 percentage points). We also analyze the performance by different categories of questions and observe that the trend of significant accuracy gains holds across most (but not all) categories, with some categories showing up to 70 percentage points improvement.

\section{Background}
\subsection{Surprisingly Common Principle in Crowdsourcing}\label{sec:related}
The problem of information elicitation without verification (i.e. when correct information is not available for scoring and when agents in a crowd need to be incentivized for providing correct information) is modelled as a game between multiple agents in the literature~\cite{faltings2022game}. One of earlier incentive mechanisms in this space is known as the Bayesian Truth Serum~\cite{prelec2004bayesian}. BTS asks every agent to submit two reports for a question. The first report is what agents believe is the correct answer to the question and in the second report, the agents predict the distribution of answers given by other agents. The reward of an agent is the sum of two reward terms. The first term (called the information score) measures the log of the ratio between the frequency of the reported answer and the geometric mean of the predictions about the answer. The second term (called prediction score) measures how close is the prediction of the agent about the distribution of other agents' answers to the actual distribution. The second term exists only to reward honest second report, but it is the first term (the information score) that is the interesting one. The resulting reward is shown to be incentive-compatible i.e. in equilibrium, agents can maximize their expected reward by telling what they believe is the correct answer.

A number of other BTS motivated reward/scoring mechanisms advanced this field of research. We focus on proposals that are suitable for crowdsourcing without requiring the agents to explicitly submit their prediction about other agents' answers. Consider for example, a crowdsourcing task of measuring pollution at locations, where no independent ground truth measurement exists. Multiple independent agents (peers) measure and report the value at a given location to a centre, but there is no trusted verification of the ground truth. Agents have to exert effort to accurately measure the value, and the centre needs to provide a reward to compensate the agents. Peer-prediction mechanisms~\cite{faltings2022game} consider this setting as a game among the agents, where each tries to maximize the expected reward attributed to their report. The  simplest form is output agreement~\cite{von2004labeling}, where reports are rewarded proportionally to the frequency of the same report among peers. However, it has been shown that the best strategies for the participating agents are always uninformative, e.g. all report the same value~\cite{jurca2007reliable}. Mechanisms, that address this issue in output agreement, are all based on the surprisingly common principle but mathematical models and game-theoretic arguments differ in different mechanisms. In a recent survey, Faltings, 2023~\cite{ijcai2023p740} identifies three types of mechanisms. The first is agreement, where the reward is proportional to the frequency of the answer among other responses, often calibrated by the overall chance of agreement~\cite{von2004labeling,dasgupta2013crowdsourced,shnayder2016informed}. The second is information-theoretic, where the reward is proportional to the pairwise mutual information between the answer and the answers given by peers~\cite{prelec2004bayesian,kong2019information,radanovic2015incentive,goel2020personalized}. The third computes the reward based on the improvement in the quality of the resulting model (the Peer Truth Serum)~\cite{radanovic2016incentives,faltings2014incentives, jurca2011incentives, faltings2017peer}.

\subsection{The Peer Truth Serum}\label{sec:related-pts}
The most relevant mechanism for this paper is the Peer Truth Serum or the PTS mechanism. In the running example of pollution measurement: if an agent $i$ reported the pollution measurement at location $l$, PTS calculates the reward for the agent as follows. PTS selects another agent $p$ (called peer) who also submitted pollution measurement for the same location $l$ (or approximately same location neighbourhood). Suppose that the agent $i$ submitted $x_i$ and the peer $p$ submitted $x_p$. Here, $x_i,x_p \in \mathcal{X}$ are pollution measurement values. For e.g., $\mathcal{X} = \{low,moderate,high,very\ high\}$. 

The reward of agent $i$ under the PTS mechanism is proportional to:
$$
\frac{\mathbbm{1}_{x_i = x_p}}{R_i(x_i)}
$$
where $R_i(x_i) = num_i(x_i)/\sum\limits_{x \in \mathcal{X}} num_i(x)$,
and $num_i(x)$ is a function that counts occurrences of $x$ in the values reported by other agents, across a large number of other locations that are a priori statistically similar. $\mathbbm{1}_{x_i = x_p}$ is indicator function ($1$ if $x_i = x_p$, $0$ otherwise).

Observe that, \emph{in expectation}, the PTS does not just reward the answer which is most likely to be given by peer, but it also scales the reward inversely by a `prior' for the answer. This prior is estimated from the answers collected from the crowd across a number of a priori statistically similar questions. This is how the PTS mechanism operationalizes the surprisingly common principle. 
In this paper, we take inspiration from the PTS mechanism to define a related notion of `surprisingly likely' textual responses of a language model.

\section{Surprisingly Likely Responses of Large Language Models}\label{sec:explanation}
We now define the notion of `surprisingly likely' responses of large language models. Consider the following example. A question ($q$) asked to an LLM is ``According to the Bible, what forbidden fruit did Adam and Eve eat in the Garden of Eden?''. A response ($r$) for this question may be  ``The Bible does not specify what kind of fruit Adam and Eve ate'' (or ``According to the Bible, Adam and Eve ate an apple'' ... etc). 

We assign a response $r$ a score $\tau(r, q)$ as follows:

\begin{equation}\label{eq:tauscorellm}
    \tau(r,q) = \frac{P(r | q)}{P(r | \text{`?'})}
\end{equation}

where $P(r | q)$ is the conditional probability of the response in the language model, given the question. $P(r | \text{`?'})$ is the conditional probability of the response in the language model, given $`?'$. We call $P(r | \text{`?'})$ as the `prior' of the response in the language model. In equation~\ref{eq:tauscorellm}, we use just a question mark $`?'$ for calculating the prior, but there might also be other possibilities. For e.g., an empty string or another reduced and similar context (examples to follow in further section). The prior might also be calculated using an average of priors obtained by conditioning on several different reduced and similar contexts. Note that marginalizing over all possible questions (question text minus `?') leads to just the probability conditioned on `?'.

We call a response $r$ surprisingly likely, if the score $\tau(r,q)$ is higher compared to other responses.

\subsection{Discussion}\label{sec:discussion-pts-llm} We now draw a comparison between the Peer Truth Serum based information elicitation discussed in Section~\ref{sec:related-pts} and LLM response generation. For answering the questions that have right and wrong answers, we can think of an LLM as modeling the frequency of occurrence of different answer strings following the question string, among all the text data used in training. We can consider each occurrence of these text snippets in training data as a separate report of the answer to the question. Reports might come from different sources of information on the Web and other sources in the training data. The score $\tau$ assigned to an answer is computed as: the probability that the same answer occurs in another text snippet following the question, divided by the probability of that answer overall in all text snippets in training. Thus, similar to the PTS reward (recall the discussion in Section~\ref{sec:related-pts}), selecting the answer with high $\tau$ can be understood as equivalent to rewarding the LLM to generate surprisingly likely answer for the context, unlike the default LLM reward of generating answer with the highest value of the numerator in the reward score $\tau$.

Some questions that naturally follow from the above discussion are: can this strategy improve the accuracy of LLM generated responses?, why?, and in which scenarios it may not work? In this paper, we take an empirical approach to address these questions. A theoretically rigorous treatment to these specific questions should be interesting future work.

As an example, consider the question: Which city in the Netherlands has the headquarters of the Dutch government? The correct answer for this question is The Hague. Suppose we used the following reduced and similar context for calculating the prior: Which city in the Netherlands has the headquarters of Z? In this case, Amsterdam could be the most likely guess (since it is the biggest city). Another reduced context could be: Which city has the headquarters of Z? In this case, probably London or New York would be the most likely guess. Similarly, we could consider: Which city? In all these cases, The Hague is very unlikely. The surprisingly likely score compensates for this low prior of The Hague compared to Amsterdam, London and New York.

This was one example of the types of questions and answers, for which we conjecture that the LLM accuracy would improve by generating surprisingly likely answers. In further sections, we analyze the promise and limitations of this approach by conducting a series of experiments with different datasets and language models. Before presenting the experimental settings and results, we summarize some most closely related work in language modeling literature.

\section{Related Work}
\subsection{PMI in Computational Linguistics}\label{sec:related-cl}
The information-theoretic measure of pointwise mutual information (PMI)~\cite{fano1961transmission} is a well-known concept in computation linguistics and natural language processing literature~\cite{jurafsky2000speech}. It has been used in use-cases such as words association~\cite{church1990word}, keyword generation improving diversity of text~\cite{mou2016sequence, yao2017towards, zhou2019unsupervised, tang2019target, takayama2019relevant}, increasing agreement with grounding~\cite{west2022probing}, abstractive summarization~\cite{van2022mutual} etc. Holtzman \textit{et al.}, 2021~\cite{holtzman2021surface} argue that since LLMs assign probability to every possible string while generating response, it creates `surface form competition' between different strings that represent the same concept. When the LLM has to make a selection from a given list of options in multiple choice questions, the correct option is not chosen because its probability mass in the LLM is shared with another similar and correct concept that may not be in the list of options to choose from. The authors showed that PMI can be helpful in that setting. Surface form competition thus is another reason that may affect the accuracy of LLMs, but it is different from the non-truthfulness problem we discussed earlier (which, for example, leads to an LLM generating answers mimicking popular human misconceptions as demonstrated by the TruthfulQA benchmark). PMI like measure is also popular in the game-theoretic information elicitation literature, albeit the definitions, the methods of measuring it and the purpose of its application are different.

\subsection{Other Closely Related Work}\label{sec:other-related-close}
Prior work~\cite{brown2020language,zhao2021calibrate} has shown that calibration techniques can help in improving accuracy in few-shot learning settings for multiple-choice question answering in language models. While they consider specific few-shot learning setting, our focus is on more general settings. We also show the results on the TruthfulQA benchmark. Further, Kumar, 2022~\cite{kumar2022answer} proposed to subtract the context-independent probability to avoid context-independent bias. This idea will be the motivation of one of the baselines in our work and we will also evaluate it on the TruthfulQA benchmark.

\section{Experiments}
\subsection{Benchmarks}\label{sec:benchmarks}
\paragraph{TruthfulQA:} The TruthfulQA benchmark~\cite{lin-etal-2022-truthfulqa} comprises 817 questions that span 38 categories, including health, law, finance and politics etc. The authors of the benchmark observed that, for questions in this benchmark, models generated many false answers that mimic popular misconceptions; in the same way as some humans would answer due to false beliefs and misconceptions. It was also observed that larger models performed worse than smaller models. Most state-of-the-art large language models continue to perform poorly on this benchmark~\cite{openai2023gpt4,touvron2023llama}. In addition to the questions, the benchmark also contains several possible answers for each question: one of the answers is marked as best answer and other answers are marked as either correct or incorrect answers. There are between 3 and 25 answers for every question in the benchmark. On average, there are 7.6 answers per question: 4.12 are incorrect, 3.47 are correct/best.

\paragraph{COPA:} The Choice Of Plausible Alternatives (COPA) benchmark~\cite{roemmele2011choice} consists of 1000 questions, split equally into development and test sets of 500 questions each. We used the development set in our experiments. Each question is composed of a premise and two alternatives, where the task is to select the alternative that more plausibly has a causal relation with the premise. The correct alternative is randomized so that the expected performance of randomly guessing is 50\%.

\paragraph{Story Cloze:} Story Cloze is a commonsense reasoning test~\cite{mostafazadeh2016corpus}; it asks a system to choose the correct ending to a four-sentence story. The benchmark contain two ending choices for each of the four-sentence story, out of which one is correct. We used the development set in our experiments which has 1871 stories.

\subsection{Measuring Conditional Probabilities in LLMs}
For our experiments with the TruthfulQA dataset, we used the logits in the pre-trained large language models for the strings `?'+$r$ and $q+r$ to obtain the cross entropy for tokens in $r$; giving us the negative of log of the conditional probabilities in the denominator and numerator respectively in equation~\ref{eq:tauscorellm} (i.e. $-\text{log }P(r|'?')$ and $-\text{log }(P(r|q))$).

For experiments with the Story Cloze benchmark, we used a similar strategy as above except that we conditioned on the last punctuation from the last input sentence of the story (instead of `?') for measuring the prior. For the COPA benchmark, we condition on `because' or `so' depending on question tag (`cause'/`effect') instead of `?' for measuring the prior. The use of last punctuation and `because' or `so' for these two benchmarks is consistent with~\cite{holtzman2021surface}, where similar idea was used (for a different reason and explanation i.e. for removing surface-form-competition; see Section~\ref{sec:related-cl}).

\subsection{Models}
We used openly available pre-trained models GPT-2 (from OpenAI)~\cite{radford2019language} and LLaMA-2 (from Meta)~\cite{touvron2023llama} in our experiments. Specifically, we used the following models: GPT-2 S (124 million parameters), GPT-2 M (355 million parameters), GPT-2 L (774 million parameters), GPT-2 XL (1558 million parameters), LLaMA-2 7B (7 billion parameters), LLaMA-2 13B (13 billion parameters) and LLaMA-2 70B (70 billion parameters). For LLaMA-2 70B, we used the 4-bit version due to compute resource constraints on our end; for all other models, we used their full precision versions. All models were obtained through Hugging Face~\cite{wolf2020transformers}\footnote{\url{https://huggingface.co/models}}. We used the publicly released \emph{base} versions of GPT-2 and LLaMA-2 for probability calculations in our experiments. Experiments were run on NVIDIA A100 with a renting cost of less than GBP 100.

\subsection{Accuracy Measure}
We measured accuracy as the fraction of questions for which the selected answer (by the respective method) was either the best answer or one of the correct answers in the benchmark.

\subsection{Baselines and Nomenclature}
For brevity, in discussion of the results, we will use the following nomenclature for various methods used in the comparison. `MaxPost' refers to a baseline standard method that selects the response with the maximum conditional probability given the question text. `MaxPostN' refers to another baseline method in which the conditional probability given the question text is normalized by the number of tokens in the response and the response with the highest normalized probability score is selected. `Top2MinPr' refers to a baseline method of shortlisting top 2 responses with highest conditional probability given the question text, and then selecting the one with the smaller prior. `Top2MaxPr' refers to the selection method of shortlisting top 2 responses with highest conditional probability given the question text, and then selecting the one with the higher prior. `Top2MaxPr' alludes to a completely opposite method i.e. selecting the unsurprisingly likely responses. 

`MaxRatio' refers to the surprisingly likely selection method introduced in Section~\ref{sec:explanation}, i.e. selecting the response with the highest ratio of the two conditional probabilities. `MaxDiff' refers to yet another baseline method that selects the response with the highest difference between the two conditional probabilities (instead of the ratio). This baseline method is motivated from the explanation by Kumar 2022~\cite{kumar2022answer}, as discussed in Section~\ref{sec:related}.

\subsection{Scope of the Experiments}
We do not use closed models such as GPT-3.5/4 in our experiments because there is lack of transparency in model development and further steps like reinforcement learning. Thus, using probability outputs (if available) from these models are not ideal for research. We note however that state-of-the-art open models (at the time of writing the paper) like LLaMA-2 are competitive in capabilities~\cite{touvron2023llama} to GPT-3.5. Further, like all commercial products in this space, closed models tend to be updated frequently, and research using such products is difficult to reproduce. In this paper, we make no claim of establishing a new state-of-the-art or beating commercial products. This work is an academic investigation, with limited compute resources, into a very specific research question introduced in Section~\ref{sec:intro}. 

We acknowledge that there are a number of new models and new benchmarks being continuously developed and released, as we write this paper. We do not believe that there is any consensus in the community regarding which benchmarks or which models are gold standard or ideal for which kind of research. We do not test all models and all benchmarks in this work, and leave this for a future endeavour with a larger research budget.

We also do not use any fine-tuned models in our experiments. The reason is that fine-tuning is a supervised approach, requiring labeled data to improve accuracy of question-answering in LLM, whereas the approach described in the paper is an unsupervised approach (i.e. it can improve the performance of base models \emph{even} when no supervision data is available for such fine-tuning).

\begin{table*}[h!]
\centering
\begin{tabular}{|l|*{6}{c|}}\hline
\multirow{2}{*}{LLM} & \multicolumn{6}{c|}{Method} \\  \cline{2-7} & MaxPost & MaxRatio & MaxDiff & MaxPostN & Top2MinPr & Top2MaxPr \\ \hline
GPT-2 S & 0.42 & 0.51 & 0.42 & 0.4 & \textbf{0.52} & 0.47\\\hline
GPT-2 M & 0.38 & \textbf{0.50} & 0.36 & 0.39 & 0.48 & 0.44\\\hline
GPT-2 L & 0.37 & \textbf{0.50} & 0.35 & 0.38 & 0.48 & 0.44\\\hline
GPT-2 XL & 0.36 & \textbf{0.52} & 0.33 & 0.36 & 0.47 & 0.43\\\hline
LLaMA-2 7B & 0.34 & \textbf{0.58} & 0.32 & 0.54 & 0.47 & 0.40\\\hline
LLaMA-2 13B & 0.43 & \textbf{0.59} & 0.45 & 0.55 & 0.54 & 0.47\\\hline
LLaMA-2 70B (4bit) & 0.37 & \textbf{0.58} & 0.36 & 0.51 & 0.50 & 0.41\\\hline
\end{tabular}
\caption{Accuracy of various methods with the 7 LLMs on the TruthfulQA Benchmark.}
\label{tab:tab1}
\end{table*}

\begin{table*}[h!]
\centering
\begin{tabular}{|l|*{3}{c|}}\hline
\multirow{3}{*}{LLM\ \ \ \ \ \ \ \ \ \ \ } &\multirow{3}{*}{\ \ \ \ \ \ Method\ \ \ \ \ \ }
&\multirow{3}{*}{Filtered Questions}&\multirow{3}{*}{Unfiltered Questions} \\ & & & \\ & & & \\\hline
\multirow{2}{*}{GPT-2 S} & MaxPost & 0.41 & 0.44\\ \cline{2-4} & MaxRatio & \textbf{0.50} & \textbf{0.53}\\ \hline
\multirow{2}{*}{GPT-2 M}  & MaxPost &  0.37 & 0.40\\ \cline{2-4} & MaxRatio & \textbf{0.46} & \textbf{0.54} \\ \hline
\multirow{2}{*}{GPT-2 L}  & MaxPost & 0.34 & 0.40 \\ \cline{2-4} & MaxRatio & \textbf{0.48}& \textbf{0.52} \\ \hline
\multirow{2}{*}{GPT-2 XL}  & MaxPost & 0.33 & 0.41\\ \cline{2-4} & MaxRatio & \textbf{0.49} & \textbf{0.55}\\ \hline
\multirow{2}{*}{LLaMA-2 7B}  & MaxPost & 0.31 & 0.38 \\ \cline{2-4} & MaxRatio & \textbf{0.56} & \textbf{0.61} \\ \hline
\multirow{2}{*}{LLaMA-2 13B}  & MaxPost & 0.43 & 0.44 \\ \cline{2-4} & MaxRatio & \textbf{0.59} & \textbf{0.59}\\ \hline
\multirow{2}{*}{LLaMA-2 70B (4bit)}  & MaxPost & 0.33 & 0.43\\ \cline{2-4} & MaxRatio & \textbf{0.59} & \textbf{0.56}\\ \hline
\end{tabular}
\caption{Accuracy on the TruthfulQA Benchmark: Separated by Adversarially Filtered vs Unfiltered Questions}
\label{tab:tab4}
\end{table*}

\subsection{Results}
\subsubsection{TruthfulQA Benchmark: Aggregate Performance Improvement}\label{sec:result-aggregate}

We first discuss the results on the TruthfulQA benchmark. Table~\ref{tab:tab1} shows the accuracy of different methods over all the questions in the TruthfulQA dataset. It is clear from the table that the surprisingly likely method beats all other baseline selection methods by significant margins. For example, the difference between the MaxPost and the MaxRatio methods is of 16 percentage points for GPT-2 XL and LLaMA-2 13B models. For the LLaMA-2 70B model, the difference is even bigger (24 percentage points). In general, the Top2MinPr method also improves results but is not as good as the MaxRatio method. MaxDiff method does not work that well and seems to reduce accuracy slightly compared to MaxPost.

Further, authors of the TruthfulQA dataset noted that the performance of language models decreased with increasing size of the models for GPT-2 and GPT-3. We observe from Table~\ref{tab:tab1} that unlike MaxPost and MaxPostN methods, the MaxRatio method is quite robust to such `inverse scaling' phenomenon.

Finally, it is also interesting to note that 4-bit quantization in LLaMA-2 70B causes a significant drop in accuracy for other methods, but MaxRatio appears quite robust to quantization as well.

\emph{Remark:} We also noticed that raising $k$ to a higher value in the Top$k$MinPr method can \textit{appear to} perform better on this dataset. A higher value of $k$ in Top$k$MinPr implies giving more weight to smaller values of the prior. A very high value of $k$ would be equivalent to almost ignoring the conditional probability given the question text. But ignoring the context almost entirely does not translate to a generally meaningful approach for response generation given a context, and it will make things worse on other kinds of benchmarks. We will discuss further in Section~\ref{sec:otherbench} that we seek an approach that not only shows better performance on TruthfulQA but the same approach should work more generally for other benchmarks too. Therefore, we report results for $k=2$ only. No such improvement in accuracy was observed for high values of $k$ for the Top$k$MaxPr method. For brevity, complete experimental data is available in supplementary material.\footnote{Instance level data is available in supplementary material, not just aggregate accuracy measures presented here. The readers can therefore also investigate the results by individual question.}
\begin{table*}[h!]
\centering
\begin{tabular}{|l|*{7}{c|}}\hline
\multirow{2}{*}{Category} & \multicolumn{2}{c|}{LLaMA-2 7B} & \multicolumn{2}{c|}{LLaMA-2 13B} & \multicolumn{2}{c|}{GPT-2 XL} \\  \cline{2-7} & MaxPost & MaxRatio & MaxPost & MaxRatio & MaxPost & MaxRatio \\ \hline
Advertising&0.38&\textbf{0.62}&\textbf{0.77}&0.69&0.62&\textbf{0.69}\\ \hline
Confusion: Other&0.00&\textbf{0.63}&0.13&\textbf{0.38}&0.00&\textbf{0.50}\\ \hline
Confusion: People&0.04&\textbf{0.74}&0.09&\textbf{0.65}&0.00&\textbf{0.61}\\ \hline
Confusion: Places&0.33&\textbf{0.93}&0.00&\textbf{0.73}&0.47&\textbf{0.67}\\ \hline
Conspiracies&0.36&\textbf{0.80}&0.60&0.60&0.40&\textbf{0.68}\\ \hline
Distraction&0.00&\textbf{0.36}&0.14&\textbf{0.29}&0.07&\textbf{0.21}\\ \hline
Economics&0.35&\textbf{0.55}&0.42&\textbf{0.58}&0.23&\textbf{0.48}\\ \hline
Education&0.00&\textbf{0.40}&0.10&\textbf{0.40}&0.20&\textbf{0.60}\\ \hline
Fiction&0.33&\textbf{0.63}&0.57&\textbf{0.73}&0.60&\textbf{0.67}\\ \hline
Finance&0.33&0.33&0.33&0.33&\textbf{0.33}&0.22\\ \hline
Health&0.25&\textbf{0.67}&0.29&\textbf{0.58}&0.24&\textbf{0.73}\\ \hline
History&0.29&\textbf{0.75}&0.42&\textbf{0.75}&0.33&\textbf{0.46}\\ \hline
Indexical Error: Identity&0.22&\textbf{0.56}&0.44&0.33&0.22&\textbf{0.44}\\ \hline
Indexical Error: Location&0.09&0.09&0.64&0.18&\textbf{0.09}&0.00\\ \hline
Indexical Error: Other&0.19&0.19&\textbf{0.76}&0.19&\textbf{0.33}&0.19\\ \hline
Indexical Error: Time&\textbf{0.44}&0.06&\textbf{0.88}&0.19&\textbf{0.50}&0.00\\ \hline
Language&\textbf{0.76}&0.67&\textbf{0.71}&0.62&\textbf{0.76}&0.52\\ \hline
Law&0.36&\textbf{0.52}&0.53&\textbf{0.64}&0.39&\textbf{0.52}\\ \hline
Logical Falsehood&\textbf{0.86}&0.29&\textbf{0.86}&0.29&\textbf{0.50}&0.14\\ \hline
Mandela Effect&\textbf{0.67}&0.50&\textbf{0.67}&0.50&\textbf{0.33}&0.17\\ \hline
Misconceptions&0.33&\textbf{0.73}&0.29&\textbf{0.70}&0.34&\textbf{0.64}\\ \hline
Misconceptions: Topical&0.25&\textbf{0.50}&0.00&\textbf{0.25}&0.00&\textbf{0.75}\\ \hline
Misinformation&\textbf{0.75}&0.08&\textbf{1.00}&0.17&\textbf{0.92}&0.17\\ \hline
Misquotations&0.50&\textbf{0.88}&0.31&\textbf{0.88}&0.13&\textbf{0.56}\\ \hline
Myths and Fairytales&0.14&\textbf{0.71}&0.19&\textbf{0.62}&0.24&\textbf{0.57}\\ \hline
Nutrition&0.25&\textbf{0.69}&0.38&\textbf{0.56}&0.31&\textbf{0.38}\\ \hline
Paranormal&0.31&\textbf{0.62}&0.27&\textbf{0.77}&0.27&\textbf{0.58}\\ \hline
Politics&\textbf{0.60}&0.10&\textbf{0.80}&0.40&\textbf{0.30}&0.00\\ \hline
Proverbs&0.11&\textbf{0.67}&0.11&\textbf{0.67}&0.28&\textbf{0.67}\\ \hline
Psychology&0.21&\textbf{0.37}&0.42&\textbf{0.47}&0.42&\textbf{0.53}\\ \hline
Religion&0.33&\textbf{0.60}&\textbf{0.47}&0.40&\textbf{0.40}&0.33\\ \hline
Science&0.11&\textbf{0.56}&0.00&\textbf{0.56}&0.00&\textbf{0.67}\\ \hline
Sociology&0.49&\textbf{0.55}&0.53&\textbf{0.56}&0.42&\textbf{0.51}\\ \hline
Statistics&\textbf{0.60}&0.40&\textbf{0.80}&0.60&\textbf{0.60}&0.20\\ \hline
Stereotypes&0.46&\textbf{0.63}&0.42&\textbf{0.83}&0.50&\textbf{0.67}\\ \hline
Subjective&0.22&\textbf{0.33}&\textbf{0.89}&0.78&\textbf{0.67}&0.44\\ \hline
Superstitions&0.50&\textbf{0.68}&0.41&\textbf{0.77}&\textbf{0.64}&0.59\\ \hline
Weather&0.53&\textbf{0.65}&0.53&\textbf{0.59}&0.53&\textbf{0.71}\\ \hline 
\end{tabular}
\caption{Accuracy on the TruthfulQA Benchmark: Separated by Question Categories}
\label{tab:tab5}
\end{table*}

\subsubsection{TruthfulQA Benchmark: Performance Improvement By Question Type}
Out of the 817 questions in the TruthfulQA benchmark, 437 are adversarially filtered questions and the rest 380 are unfiltered questions. The adversarially filtered questions were the ones that the authors of TruthfulQA selected based on the observed pattern of LLM producing wrong answers for them. The unfiltered questions did not go through similar filtering but they too were crafted based on the expectation that LLMs would produce wrong answers for them. Table~\ref{tab:tab4} shows the comparison of MaxPost and MaxRatio methods for the unfiltered and filtered questions using different LLMs. We observe from the table that the aggregate gain in accuracy for MaxRatio over MaxPost that we saw earlier comes from both types of questions. We also observe that there is generally a trend that adversarially filtered questions contribute slightly more gain in accuracy than unfiltered.

\subsubsection{TruthfulQA Benchmark: Performance Improvement By Answer Type}
We also investigated how many questions that were correctly answered by MaxPost but incorrectly by MaxRatio and how many questions that were correctly answered by both MaxPost and MaxRatio. The motivation for this error analysis is to confirm that the accuracy gain for MaxRatio is not due to simply selecting an opposite answer than MaxPost. For LLaMA-2 7B, we observed that for 313 questions MaxPost was wrong but MaxRatio was correct, and for 161 questions both gave correct answers. In contrast, for 120 questions, MaxPost gave correct answers but MaxRatio gave incorrect answers. For the remaining 223 questions, both gave incorrect answers. For brevity, we do not report these numbers for other models in the paper, but instance level data is available in supplementary material.

\subsubsection{TruthfulQA Benchmark: Performance Improvement By Question Category}\label{sec:byquestioncat}
The questions in the TruthfulQA are also categorised by labels such as economics, education, health, distraction, misconceptions, conspiracies, history, misinformation, politics, nutrition etc. There are 38 diverse categories in the benchmark. Table~\ref{tab:tab5} shows the breakdown of performance of MaxPost and MaxRatio by question categories. For brevity, we report this breakdown for GPT-2 XL, LLaMA-2 7B and LLaMA-2 13B. We observe from the table that for majority of the question categories, MaxRatio performs better than MaxPost. For example, in the case of LLaMA-2 7B: in 28 categories MaxRatio works better, in 7 categories MaxPost works better, and in 3 categories both methods perform same. This shows that the improvement by MaxRatio is not due to a few types of questions only but can be observed across the wide spectrum of questions. In some categories, there is an improvement of up to 70 percentage points. We do acknowledge however that these numbers get less conclusive since the benchmark is now divided into 28 smaller categories and it would be useful in the future to find bigger benchmarks on each of these categories and evaluate the methods on each benchmark.

We did not find an obvious pattern about the types of questions where MaxPost works better than MaxRatio but a few observations are worth noting. For example, the category Logical Falsehood is about tautology, like ``Are all real numbers real numbers?''. MaxRatio does not work well on these kinds of questions. Considering the motivation for the surprisingly likely method, we did not expect the method to improve accuracy on such questions, but the fact that accuracy decreased is a negative result. Further, we also noted that for many of the categories where MaxPost does better (e.g., for Indexical Error: Time and Misinformation categories, that have significant drop), the correct and best answers in the benchmark is just ``I have no comment.''. We conjecture that it may be possible to handle these categories of questions (or answers) based on a hybrid method that also uses a minimum threshold for conditional probability in the numerator or for the ratio. For example, ``No, all real numbers are not real numbers?'' has a low conditional probability given the question text ``Are all real numbers real numbers?''. This threshold would become a hyper-parameter of the method and would need to be tuned to maximize the accuracy across different tasks and question categories. Similarly, an uninformative answer ``I have no comment.'' can be encouraged if the conditional probability in the numerator or the ratio is not high enough for possible generations. It would be interesting to investigate this further in future work.

\subsubsection{COPA and StoryCloze Benchmarks}\label{sec:otherbench}
The results on the TruthfulQA benchmark show that the surprisingly likely method does help with the non-truthfulness problem in LLMs in most categories of questions. We next also test the methods on two other benchmarks (COPA and StoryCloze) to show that the surprisingly likely method, at least, does not make things worse on these other benchmarks. This test is important because TruthfulQA is a somewhat special benchmark (it contains questions that LLMs are more likely to get wrong than to get right). While it is easy to develop methods that \textit{appear to} work well only on such special benchmarks (for example, by simply flipping the answers), it is difficult to design general methods that work well on special benchmarks without degrading performance on others benchmarks. Due to the constraints on computational resources, we can not perform exhaustive testing on all other benchmarks. However, by testing on COPA and StoryCloze, we conduct a preliminary investigation in that direction.

COPA and StoryCloze benchmarks have only two choices in the dataset. We do not report Top$2$MinPr and Top$2$MaxPr for these benchmarks because that would be equivalent to ignoring the context and choosing an answer only based on prior of the answers, which does not translate to a generally meaningful approach for response generation given a context (as was also discussed in Section~\ref{sec:result-aggregate} for high values of $k$ in the TruthfulQA benchmark). We observe from Tables~\ref{tab:tab6} and~\ref{tab:tab7} that the surprisingly likely criterion either improves the performance or in a few cases leaves the performance unchanged\footnote{There is an unexplained observation in Tables~\ref{tab:tab6} and~\ref{tab:tab7}: for LLaMA-2 13 B, all methods perform relatively bad. We double-checked our code and experiment data and also re-ran the experiments/calculations, but the reason of this anomaly is not clear.}. This provides preliminary evidence that the surprisingly likely method does offer benefit on the TruthfulQA benchmark without decreasing the accuracy on other benchmarks.

\begin{table*}[h!]
\centering
\begin{tabular}{|l|c|c|c|c|}\hline
\multirow{2}{*}{LLM} & \multicolumn{4}{c|}{Method} \\
\cline{2-5} & MaxPost & MaxRatio & MaxDiff
& MaxPostN \\\hline
GPT-2 S & 0.61 & \textbf{0.63} & 0.62 & \textbf{0.63} \\\hline
GPT-2 M & 0.67 & \textbf{0.70} & 0.67 & 0.66 \\\hline
GPT-2 L & \textbf{0.70} & 0.69 & \textbf{0.70} & 0.68 \\\hline
GPT-2 XL & 0.69 & \textbf{0.72} & 0.69 & 0.68 \\\hline

LLaMA-2 7B & 0.82 & \textbf{0.83} & 0.82 & 0.69  \\ \hline

LLaMA-2 13B & 0.61 & \textbf{0.65} & 0.49 & 0.51 \\ \hline

LLaMA-2 70B (4bit) & \textbf{0.88} & \textbf{0.88} & 0.87 & 0.74 \\ \hline
\end{tabular}
\caption{Results on the COPA Benchmark}
\label{tab:tab6}
\end{table*}

\begin{table*}[h!]
\centering
\begin{tabular}{|l|c|c|c|c|}\hline
\multirow{2}{*}{LLM} & \multicolumn{4}{c|}{Method} \\
\cline{2-5} & MaxPost & MaxRatio & MaxDiff
& MaxPostN \\\hline
GPT-2 S & 0.58 & \textbf{0.67} & 0.58 & 0.60 \\\hline
GPT-2 M & 0.62 & \textbf{0.71} & 0.62 & 0.67 \\\hline
GPT-2 L & 0.64 & \textbf{0.72} & 0.64 & 0.69 \\\hline
GPT-2 XL & 0.67 & \textbf{0.76} & 0.67 & 0.72 \\\hline

LLaMA-2 7B & 0.77 & \textbf{0.82} & 0.69 & 0.68  \\ \hline

LLaMA-2 13B & 0.54 & \textbf{0.63} & 0.52 & 0.53 \\ \hline

LLaMA-2 70B (4bit) & 0.77 & \textbf{0.85} & 0.68 & 0.70  \\ \hline
\end{tabular}
\caption{Results on the StoryCloze Benchmark}
\label{tab:tab7}
\end{table*}

\section{Limitations}\label{sec:limitations}
The notions of `truth' and `truthfulness' are difficult to formalize and there is often much philosophical debate about these terms. In this paper, we restricted our discussion to questions in which it is reasonable to assume that there exist objectively correct and incorrect responses. Further, we also assume that given a sufficiently clear prompt, the desired behaviour of LLMs is to generate a correct response. For example, consider the question, ``Which city is the capital of Brazil?''. We assume that the desired behaviour of LLM for this clearly written prompt is not to generate ``São Paulo'' or ``Rio de Janeiro''; instead it is to generate ``Brasília''. We hypothesized that, besides other possible reasons, LLMs may produce incorrect response (i.e. be non-truthful) for such questions due to mis-specified (non-truthful) training objective, or due to incorrect or sub-optimal aggregation of information in its noisy training data. 

In this paper, we did not delve into the discussion on subjective information like opinions or beliefs. Inherently subjective information can not be categorized as correct or incorrect in the same way as objective information; a possible ground truth in such cases is perhaps the underlying distribution of subjective opinions or beliefs across the specified population. Further, in such cases, truthful behaviour of an agent is generally defined as answering honestly or not lying about its opinions and beliefs. These notions of truth and truthfulness are included in the broader truthful information elicitation literature, but were not discussed for LLMs in our work. The reason we did not discuss these is that the interpretation of terms like  opinions, beliefs and honesty in the case of LLMs is not quite the same as in the case of agents in a crowd. Separate careful discussion is required to understand when it makes sense to responsibly use these terms in the case of LLMs and what these terms mean precisely in given context. We leave this discussion for future work.

\section{Conclusions and Future Work}\label{sec:future}
In this paper, we defined the notion of ‘surprisingly likely’ textual response of a large language model. This
notion was inspired by the surprisingly common principle in the crowdsourcing literature, but tailored for text in a language model. We observed that surprisingly likely responses of
large language models are more accurate on the TruthfulQA benchmark compared to baselines. We also discussed the strengths and limitations of this approach by analysing performance across different types of questions/answers. This work is one of the early attempts to bridge two different research fields of language modeling and collective intelligence systems, and we hope that it will motivate further research at the intersection of the two research fields.

For example, an interesting future work that may directly follow this work is to construct theoretical models to provably explain the observations on different categories of questions and further understand the strengths and weakness of the approach.
Finally, while our experiments show that the surprisingly likely responses are indeed more correct, it remains future work to show how this can be best implemented to make LLMs generate more correct responses in the first place. It would also allow researcher to obtain results on benchmarks other than multiple choice questions benchmarks (e.g., open-ended questions benchmarks). Possible ideas include intervening at decoding stage or at pre-training or later stages through reinforcement learning.

\section{Acknowledgements}
The author was supported by Oxford Martin's programme on `Ethical Web and Data Architectures (EWADA) in the Age of AI'. Special thanks to Prof Boi Faltings for his participation in many discussions that significantly helped the author while writing the paper. The author also thanks Dr. Debjit Paul for his kind help in running an earlier version of our code on a compute cluster. Any errors in the paper are of the author only.

\bibliographystyle{unsrturl}
\bibliography{truthful_llm}

\end{document}